\pdfoutput=1

\documentclass[11pt]{article}

\usepackage[]{acl}

\usepackage{times}
\usepackage{latexsym}

\usepackage[T1]{fontenc}

\usepackage[utf8]{inputenc}

\usepackage{microtype}

\usepackage{inconsolata}

\usepackage{graphicx}

\usepackage[table]{xcolor}
\usepackage{booktabs}   
\usepackage{colortbl}   
\usepackage{siunitx}
\usepackage{graphicx}
\usepackage{array}
\usepackage{caption}
\usepackage{subcaption} 
\usepackage{multirow}

\title{Evaluating the Creativity of LLMs in Persian Literary Text Generation}

\author{
    \textbf{Armin Tourajmehr*$^1$} ~
    \textbf{Mohammad Reza Modarres*$^1$} ~
    \textbf{Yadollah Yaghoobzadeh$^{2,1}$} \\
    $^1$Tehran Institute for Advanced Studies, Khatam University, Iran \\
    $^2$School of Electrical and Computer Engineering, \\College of Engineering, University of Tehran, Tehran, Iran\\
    \texttt{a.tourajmehr@teias.institute} ~
    \texttt{mr.modarres@teias.institute} ~
    \texttt{y.yaghoobzadeh@ut.ac.ir}
}

\begin{document}
\maketitle
\begin{abstract}
\begingroup
  \hypersetup{hidelinks}       
  \renewcommand{\thefootnote}{*} 
  \footnotetext{Equal contribution}
\endgroup

Large language models (LLMs) have demonstrated notable creative abilities in generating literary texts, including poetry and short stories. However, prior research has primarily centered on English, with limited exploration of non-English literary traditions and without standardized methods for assessing creativity. In this paper, we evaluate the capacity of LLMs to generate Persian literary text enriched with culturally relevant expressions. We build a dataset of user-generated Persian literary spanning 20 diverse topics and assess model outputs along four creativity dimensions—originality, fluency, flexibility, and elaboration—by adapting the Torrance Tests of Creative Thinking. To reduce evaluation costs, we adopt an LLM as a judge for automated scoring and validate its reliability against human judgments using intraclass correlation coefficients, observing strong agreement. In addition, we analyze the models’ ability to understand and employ four core literary devices: simile, metaphor, hyperbole, and antithesis. Our results highlight both the strengths and limitations of LLMs in Persian literary text generation, underscoring the need for further refinement.\footnote{The dataset, code, and evaluation guide are available at \href{https://github.com/Armin-Tourajmehr/Evaluating-the-Creativity-of-LLMs-in-Persian-Literary-Text-Generation}{github}}
\end{abstract}

\section{Introduction}

As LLMs continue to evolve and gain widespread use, there has been growing interest in their potential to perform tasks that require creativity. A prominent application is creative writing, where LLMs are increasingly employed to generate stories, poetry, and other literary forms. Yet debate persists over whether these models can genuinely emulate or replace human writers in producing creative text \citep{gervais2024creative}. A common criticism is that LLMs struggle with creativity, particularly in generating original, high-quality, and culturally nuanced outputs \cite{boussioux2024crowdless, chakrabarty2023art, gomez-rodriguez2023confederacy}.

Most existing research on creative text generation has focused on English and English-speaking contexts. Consequently, the creative capabilities of LLMs in other languages—especially low-resource ones such as Persian—remain largely underexplored. Despite the significance of literary text generation as a distinct form of creative expression, to our knowledge no study has systematically evaluated LLM-generated literary texts in Persian. Moreover, prior work—whether in Persian or English—rarely considers the challenge of evaluating culturally grounded literary texts produced by native speakers, beyond limited domains such as story writing and poetry. This gap underscores the need for evaluating how well models align with the cultural and literary practices of human communities.

Evaluating creativity in LLMs presents unique challenges due to their distinct reasoning processes, the subjective nature of creativity, and the limitations of manual evaluation. A widely used framework for assessing human creativity is the Torrance Tests of Creative Thinking (TTCT) \cite{torrance1966ttct}, which evaluate four core dimensions: originality, fluency, flexibility, and elaboration. Current benchmarks such as the Alternative Uses Task (AUT) \cite{stevenson2022gpt3, summersstay2023brainstorm} capture divergent thinking but fall short in addressing the cultural depth and stylistic richness required for literary creativity.

Building on the work of \citet{zhao2024assessing}, who adapted the TTCT for evaluating general-purpose creativity in LLMs, we extend this approach to Persian literary text generation—a domain that poses unique linguistic and cultural challenges. Unlike prior studies that focus on open-ended prompts, our framework emphasizes the generation of stylistically rich and culturally grounded Persian sentences. To this end, we introduce a culturally adapted evaluation framework based on the four TTCT dimensions. Moreover, in the absence of suitable resources, we compile and release \textit{CPers} (Creativity in Persian)—the first dataset designed for this purpose—which provides a foundation for systematic creativity evaluation in low- and mid-resource languages such as Persian.

Evaluating six state-of-the-art LLMs—including GPT-3.5, GPT-4.1, DeepSeek-V3, DeepSeek-R1, Qwen2.5, and Gemma—we conduct one of the first systematic analyses of Persian literary creativity in LLM outputs. To ensure reliable scoring, we combine human annotations with an LLM-as-a-judge framework using Claude 3.7 Sonnet, which demonstrates strong alignment with human evaluators. Beyond creativity evaluation, we conduct two complementary studies: first, we examine word usage across topics to assess each model’s cultural alignment with native Persian speakers and values; second, we analyze the presence of four literary devices frequently used in Persian literature—simile, metaphor, antithesis, and hyperbole—to explore how stylistic elements relate to creativity.

Our work makes the following key contributions:

\begin{itemize}
    \item We present the first systematic evaluation of LLMs on Persian literary text generation, adapting the Torrance Tests of Creative Thinking (TTCT) to assess originality, fluency, flexibility, and elaboration in a culturally grounded context.  

    \item We introduce \textit{CPers}, a novel dataset of 4,371 Persian literary texts spanning 20 emotionally and culturally diverse topics, authored by native speakers, along with a human-annotated subset of 200 texts that includes creativity scores and labels for rhetorical devices. This dataset provides a benchmark for evaluating creativity in low-resource languages.  

    \item We conduct a comprehensive evaluation of six state-of-the-art LLMs, combining human annotations and automated scoring via Claude 3.7 Sonnet, and analyze their performance across multiple dimensions of creativity.  

    \item We investigate the use of key literary devices—simile, metaphor, antithesis, and hyperbole—as well as lexical patterns, revealing how the balance and nuanced deployment of these devices influence perceived creativity and cultural alignment.  

    \item Our analysis provides insights into model design and training strategies, showing, for example, that reasoning-oriented models produce more elaborated and flexible literary texts, and highlighting the importance of multi-dimensional, culturally aware evaluation for creative text generation.
\end{itemize}

\begin{table*}[t]
\small
\centering
\begin{tabular}{ |p{3 cm}| p{12 cm}| }
 \hline
 \multicolumn{1}{|c|}{\textbf{Criteria}} & \multicolumn{1}{c|}{\textbf{Questions}} \\
 \hline
 \begin{center}\textbf{Originality}\end{center} &- Is the sentence creative and far from common clichés? \newline
- Is the sentence similar to famous sentences, poems, or Persian proverbs? \newline
- Does the sentence contain at least one of the literary devices of simile, metaphor, antithesis, or hyperbole? \\
 \hline
 \begin{center}\textbf{Fluency}\end{center} & - Is the sentence grammatically correct and understandable? \newline
- Does the sentence seem fluent and natural to a Persian reader? \newline
- Can the sentence be used in a literary text or everyday conversation? \\
 \hline
 \begin{center}\textbf{Flexibility}\end{center} & - Does the sentence use multiple ideas to express the intended topic? \newline
- Does the sentence look at the topic from a new perspective? \newline
- Does the sentence use different styles (e.g., ironic, humorous, philosophical)? \\
 \hline
 \begin{center}\textbf{Elaboration}\end{center} & - Does the sentence go into detail and use a variety of vocabulary? \newline
- Does the sentence create a clear mental image in the reader? \newline
- Does the sentence convey a specific feeling (e.g., love, sadness, hope) well? \\
 \hline
\end{tabular}

\caption{A creativity assessment framework for Persian texts, grounded in originality, fluency, flexibility, and elaboration.}
\label{tab:criteria_questions}
\end{table*}

\section{Related Work}
Creative writing is a cognitively complex and performative language task that requires linguistic fluency, cultural and literary competence, narrative coherence, and the capacity for originality and imagination. Recent work has increasingly explored the use of LLMs in creative domains, including humor generation \cite{zhong2023lets}, comedy creation \cite{mirowski2024robot}, and psychological creativity assessments \cite{bellemarepepin2024divergent}. Studies show that LLMs can produce poetic and narrative content of high quality \cite{franceschelli2023creativity}, and that human judges often struggle to distinguish between human-written and model-generated stories \cite{clark2021all}.

Creativity, however, remains difficult to evaluate due to its subjective nature. Common assessment methods include the Divergent Association Task (DAT) \cite{olson2021naming}, the Remote Associates Test (RAT) \cite{mednick1962associative}, and the widely used Torrance Tests of Creative Thinking (TTCT) \cite{torrance1966ttct} in psychometric studies.

\citet{stevenson2022gpt3} show that human creative outputs outperform those of GPT-3 on the Alternative Uses Task. \citet{summersstay2023brainstorm} further demonstrate that while GPT-3 could generate original ideas, it often failed to filter out impractical ones. \citet{naeini2023redherrings} introduce the OnlyConnect Wall dataset to simulate RAT-like tasks for evaluating creative problem solving in LLMs. Their findings reveal that red herrings reduce model performance, though their analysis does not incorporate advanced prompting or retrieval-augmented methods, leaving room for further exploration. Similarly, \citet{atmakuru2024cs4} propose the CS4 benchmark to assess creativity under varying prompt specificity, promoting originality over memorization. Unlike CS4’s focus on general storytelling, our work specifically targets literary creativity, with emphasis on style, emotion, and cultural depth.

Recent narrative-level analyses \cite{tian2024are} show that LLMs systematically generate stories that are more predictable and positive, while struggling with managing climaxes and emotional arcs. Structural approaches such as marking turning points can improve narrative quality, but a substantial gap with human writing remains. Other studies indicate that, although LLMs perform strongly in linguistic fluency, they still lag behind humans in novelty, diversity, and surprise \cite{ismayilzada2025evaluating}. Collaborative generation with multiple models can enhance diversity and creativity, but often at the cost of coherence \cite{venkatraman2025collabstory}.

\citet{zhao2024assessing} present a scalable benchmark for evaluating LLM creativity using a modified version of the TTCT and automated GPT-4 scoring on general creative tasks. While their work demonstrates the feasibility of large-scale creativity testing in English, it does not account for cultural or linguistic differences that are central to literary creativity.

In contrast, our study is the first to evaluate LLM-generated literary creativity in Persian, a culturally rich yet underrepresented language. We adapt the TTCT to reflect stylistic, emotional, and metaphorical aspects characteristic of Persian literature. This not only fills an important gap in cross-lingual creativity evaluation but also offers a framework for studying literary creativity in diverse cultural traditions, with potential extensions to other languages.

\begin{figure}[ht]
    \centering
    \includegraphics[width=0.48\textwidth]{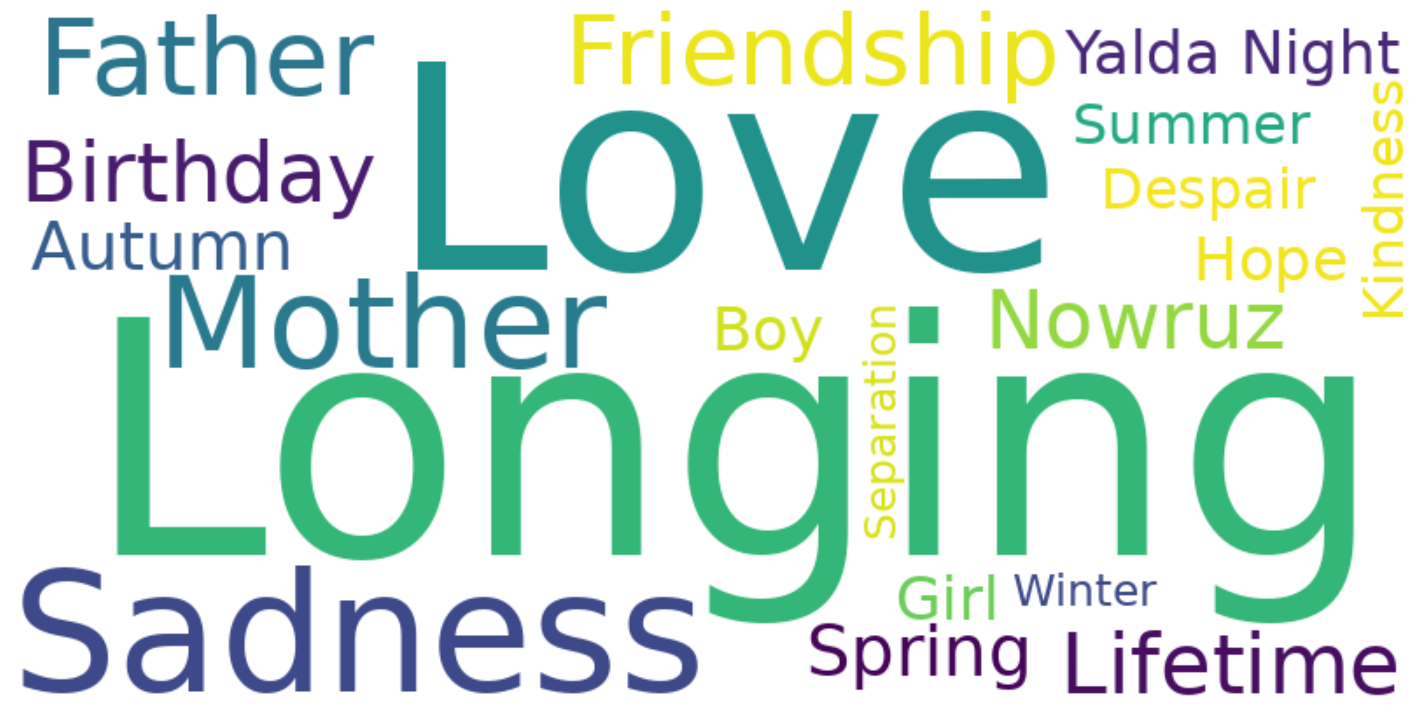}
    \caption{Word cloud illustrating the distribution of themes in the CPers dataset. The most frequent theme appears 670 times, while the least frequent occurs 50 times.}
    \label{fig:word-claude}
\end{figure}

\section{Methodology}
We propose a framework for evaluating the creativity of LLMs in Persian literary text generation. Our evaluation builds on the Torrance Tests of Creative Thinking (TTCT) by focusing on four dimensions: originality, fluency, flexibility, and elaboration. We construct a dataset of Persian literary texts and employ both human annotators and LLM-based reviewers. 

\subsection{CPers Dataset}
\label{sec:dataset}
\begin{figure}[ht]
    \centering
    \includegraphics[width=0.48\textwidth]{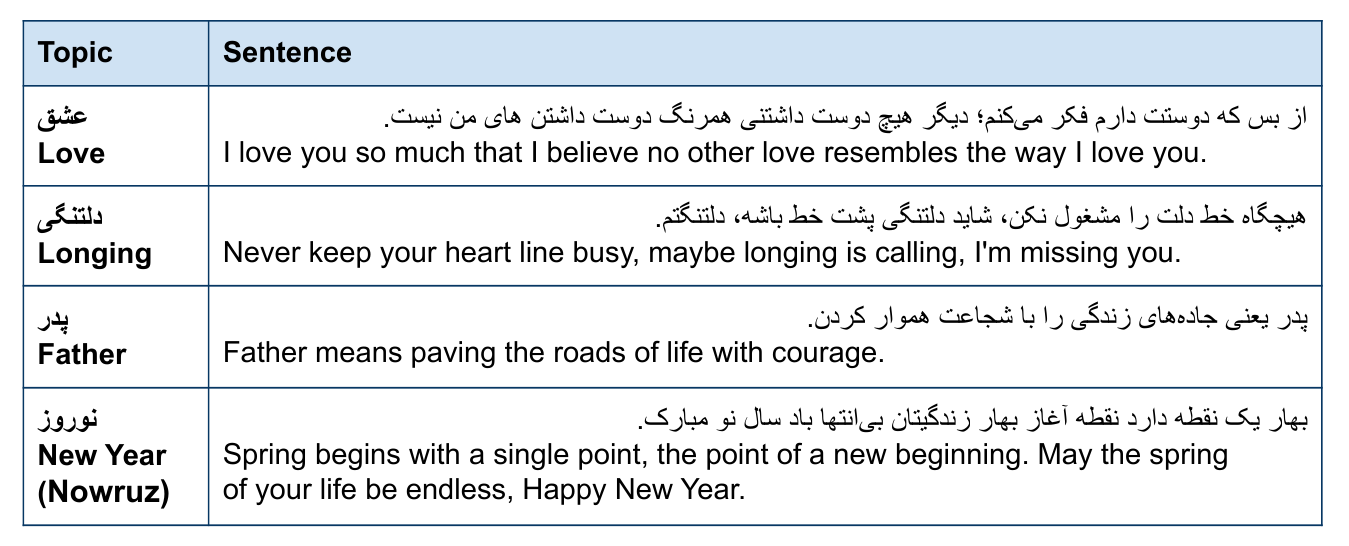} 
    \caption{Sample entries from the \textit{CPers} dataset, showing Persian literary sentences and their emotional or cultural topics, along with English translations.
    Note that Nowruz, the Persian New Year, coincides with the first day of spring.}
    \label{fig:creapers-table}
\end{figure}

To conduct this study, access to a dataset specifically tailored to Persian literary texts is essential. However, no publicly available dataset rooted in Persian-speaking culture exists. We therefore create a new resource, \textit{CPers}, by collecting texts from various online sources. The final dataset comprises 4,371 texts spanning 20 distinct topics. Authored by everyday people, these writings capture a wide range of human emotions and relationships. Each text averages approximately 26 words in length. The distribution of topics is balanced, with no category exceeding 15\% or falling below 1\% of the corpus (see Figure~\ref{fig:word-claude}). This balance ensures diverse coverage of cultural and emotional themes. Additional details on dataset construction and a full list of topics are provided in Appendix~\ref{sec:appendix}. Representative samples are shown in Figure~\ref{fig:creapers-table}.

The topics include universal themes such as love, kindness, hope, disappointment, friendship, and sadness, as well as culturally significant occasions such as Nowruz (Persian New Year), Father's Day, and Mother's Day. Although most texts were produced by non-professional writers or inspired by classical literary figures, efforts were made to preserve literary richness in the majority of cases. During dataset construction, we prioritized texts that incorporated at least one rhetorical device—such as simile, metaphor, antithesis, or hyperbole—so that the collection would reflect not only everyday language use but also the stylistic depth characteristic of Persian literary expression.

\begin{figure*}[t!]
    \centering
    \subfloat[ \label{subfig1:a}]{
    \includegraphics[height=.35\textwidth,trim=10mm 0 10mm 0]{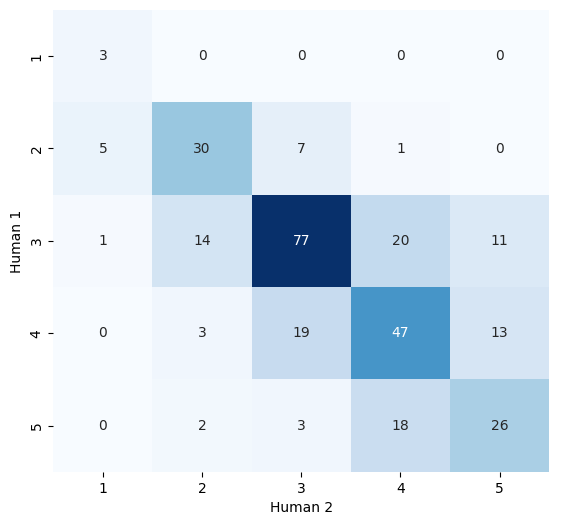}
    
    }\hspace{5mm}
    \subfloat[ \label{subfig1:b}]{
	\includegraphics[height=.35\textwidth,trim=10mm 0 10mm 0]{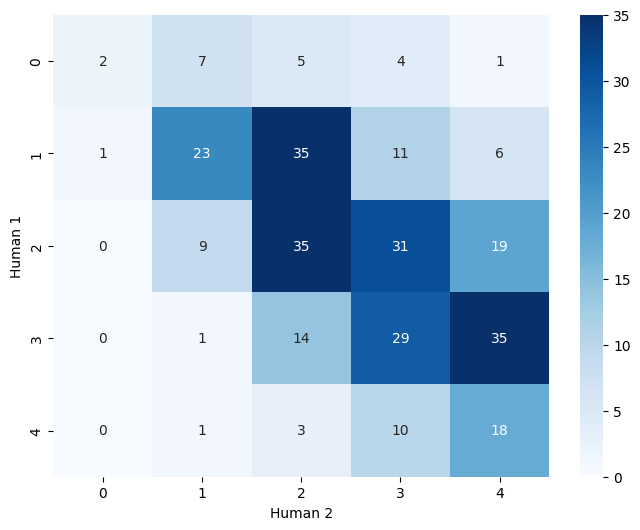}
	}
	\vspace{-3mm}
    \caption{Confusion matrices showing inter-rater agreement between two human annotators for the originality criterion on: (a)  Model texts, and (b) Human texts, within the proposed framework.
    }
    \label{fig:dataset-level}
    \vspace{-3mm}
\end{figure*}

\subsection{Evaluation Metric}
\label{sec:metric}
\label{sec:ttct}

To assess the creativity of texts, whether human- or LLM-generated, we develop a new evaluation framework inspired by the TTCT and specifically adapted to the Persian-speaking cultural context. The framework is organized around four key dimensions—originality, fluency, flexibility, and elaboration. Each dimension is assessed through three culturally tailored questions, yielding a total of 12 questions (see Table \ref{tab:criteria_questions}). Responses are rated on a five-point scale, where 1 indicates the lowest and 5 the highest score. This structure enables a systematic and culturally relevant assessment of creative text generation.

A central adaptation concerns the notion of \emph{fluency}. In the original TTCT, fluency is often measured quantitatively as ideational fluency—the number of distinct ideas produced. This metric, however, is not directly applicable to short Persian texts, particularly those generated by LLMs. Creative ideas in Persian are frequently conveyed implicitly or metaphorically within a single sentence, making idea-counting both ambiguous and culturally biased. To address this, we redefine fluency to evaluate the grammatical accuracy of the text, its naturalness for native speakers, and its appropriateness for either literary or conversational contexts. This redefinition aligns more closely with how fluency is perceived in Persian creative writing. The questions are developed through iterative refinement, informed by pilot annotation sessions and error analyses of both human-written and model-generated Persian texts.

The decision to adapt the TTCT for Persian arises from both linguistic and cultural considerations. These adaptations are necessary to ensure construct validity and cultural fairness in evaluating creativity. Although our implementation is tailored for Persian, the overall structure and methodology are general and can be extended to other languages and cultural contexts with appropriate modifications.

\begin{table}[t]
\centering
\resizebox{\columnwidth}{!}{%
\rowcolors{2}{gray!15}{white}
\begin{tabular}{lcccc}
\toprule
\textbf{Text type} & \textbf{Originality} & \textbf{Fluency} & \textbf{Flexibility} & \textbf{Elaboration} \\
\midrule
Model text  & 0.70 & 0.78 & 0.76 & 0.69 \\
Human text  & 0.67 & 0.42 & 0.45 & 0.69 \\
\bottomrule
\end{tabular}
}
\caption{\label{table:eval-humans}ICC scores of Human-1 with Human-2 on \textit{Model texts} and \textit{Human texts} (p-value $<< 0.05$ for all dimensions).}
\end{table}

\begin{table*}[t]
\centering

\begin{minipage}{0.49\textwidth}
\centering
\resizebox{\textwidth}{!}{%
\rowcolors{2}{gray!15}{white}
\begin{tabular}{lcccc}
\toprule
\textbf{Model} & \textbf{Originality} & \textbf{Fluency} & \textbf{Flexibility} & \textbf{Elaboration} \\
\midrule
GPT-4o            & 0.57 & 0.63 & 0.67 & 0.30 \\
Claude 3.7 Sonnet & 0.46 & 0.69 & 0.55 & 0.54 \\
\bottomrule
\end{tabular}
}
\subcaption{\label{table:eval-a}ICC of the average scores of Human-1 and Human-2 on \textit{Model texts} (p-value $<< 0.05$ for all dimensions).}
\end{minipage}
\hfill
\begin{minipage}{0.49\textwidth}
\centering
\resizebox{\textwidth}{!}{%
\rowcolors{2}{gray!15}{white}
\begin{tabular}{lcccc}
\toprule
\textbf{Model} & \textbf{Originality} & \textbf{Fluency} & \textbf{Flexibility} & \textbf{Elaboration} \\
\midrule
GPT-4o            & 0.64 & -0.26 & 0.61 & 0.58 \\
Claude 3.7 Sonnet & 0.65 & 0.39 & 0.46 & 0.59 \\
\bottomrule
\end{tabular}
}
\subcaption{\label{table:eval-b}ICC of the average scores of Human-1 and Human-2 on \textit{Human texts} (p-value $<< 0.05$ for all dimensions).}
\end{minipage}

\caption{\label{table:eval-models}ICC of the average scores of Human-1 and Human-2 with models across four TTCT-based dimensions.}
\end{table*}

\subsection{Human Annotated Dataset} \label{sec:human_set}

One hundred instances are selected from the \textit{CPers} dataset, referred to as \textit{Human texts}. These texts cover five topics—\textit{love}, \textit{longing}, \textit{friendship}, \textit{hope}, and \textit{despair}—representing a balanced range of human emotions. Using GPT-3.5 \cite{openai2023gpt35} as the base model, an additional 100 literary texts are generated across the same five topics via zero-shot prompting (as described in \ref{sec:setup}), referred to as \textit{Model texts}, ensuring alignment with the human-written topics.

To establish a ground truth for creativity evaluation, two human annotators assessed both 100 \textit{Human texts} and 100 \textit{Model texts} using the proposed 12-question framework covering the four key dimensions. For each text, an overall creativity score was computed by averaging the scores across these dimensions.

To ensure consistency among annotators, calibration meetings were conducted using a demo dataset prior to the main annotation task. Inter-rater reliability was assessed by examining the variation in scores assigned by different annotators. The confusion matrix illustrating agreement on the originality criterion is shown in Figure~\ref{fig:dataset-level}\footnote{For results on other criteria, see Appendix \ref{appendix:conf}.}. As observed, in most cases, ratings differed by no more than one point.

To quantify agreement, we employ the Intraclass Correlation Coefficient (ICC), a statistical measure that evaluates the consistency of observations within groups. Unlike Pearson correlation, which only captures linear relationships, ICC assesses the closeness of scores, making it particularly suitable in our context where most inter-rater differences fall within a single point on the 1–5 scale. ICC therefore provides a more appropriate measure of agreement for two raters independently scoring each sentence. The ICC values for \textit{Human texts} and \textit{Model texts} are presented in Table~\ref{table:eval-humans}. These results indicate strong and consistent agreement across both text types and all creativity dimensions, reflecting the inherently subjective nature of evaluating creative writing \citep{gomez-rodriguez2023confederacy}.
Notably, as shown in Table~\ref{table:eval-humans}, annotators exhibit slightly higher agreement on model-generated texts across all evaluation criteria.

\subsection{LLM as Judge}\label{sec:judge}
To establish a robust and consistent framework for evaluating the creativity of LLM-generated texts, we compare the judgment behavior of two general-purpose language models—GPT-4o \cite{openai2024gpt4o} and Claude 3.7 Sonnet \cite{anthropic2025claude37}. The primary criterion for selecting a model as a judge is its alignment with human evaluators, since creativity—particularly in literary and culturally nuanced contexts—is inherently subjective and challenging to assess automatically.

As reported in Table~\ref{table:eval-humans}, ICC scores between the two human annotators are consistently high across all four TTCT dimensions, indicating strong inter-annotator reliability. We therefore use the average of their scores as the human gold standard and compute ICC values between these averages and the ratings produced by each LLM.

Quantitative results indicate that Claude 3.7 Sonnet exhibits stronger alignment with human annotations compared to GPT-4o. As shown in Tables~\ref{table:eval-a} and \ref{table:eval-b}, Claude’s creativity ratings across the four TTCT dimensions—originality, fluency, flexibility, and elaboration—closely match those of human annotators for both \textit{Human texts} and \textit{Model texts}. Claude achieves high average ICC scores with both human raters across all criteria and text types, whereas GPT-4o shows lower average correlations and even a negative correlation in one instance.

We also observe that GPT-4o tends to assign higher scores to \textit{Model texts} than to \textit{Human texts}, introducing a bias that complicates the assessment of literary creativity and limits deeper analysis of stylistic qualities.

\begin{figure}[h]
    \centering
    \includegraphics[width=0.48\textwidth, trim=0 0 0 0, clip]{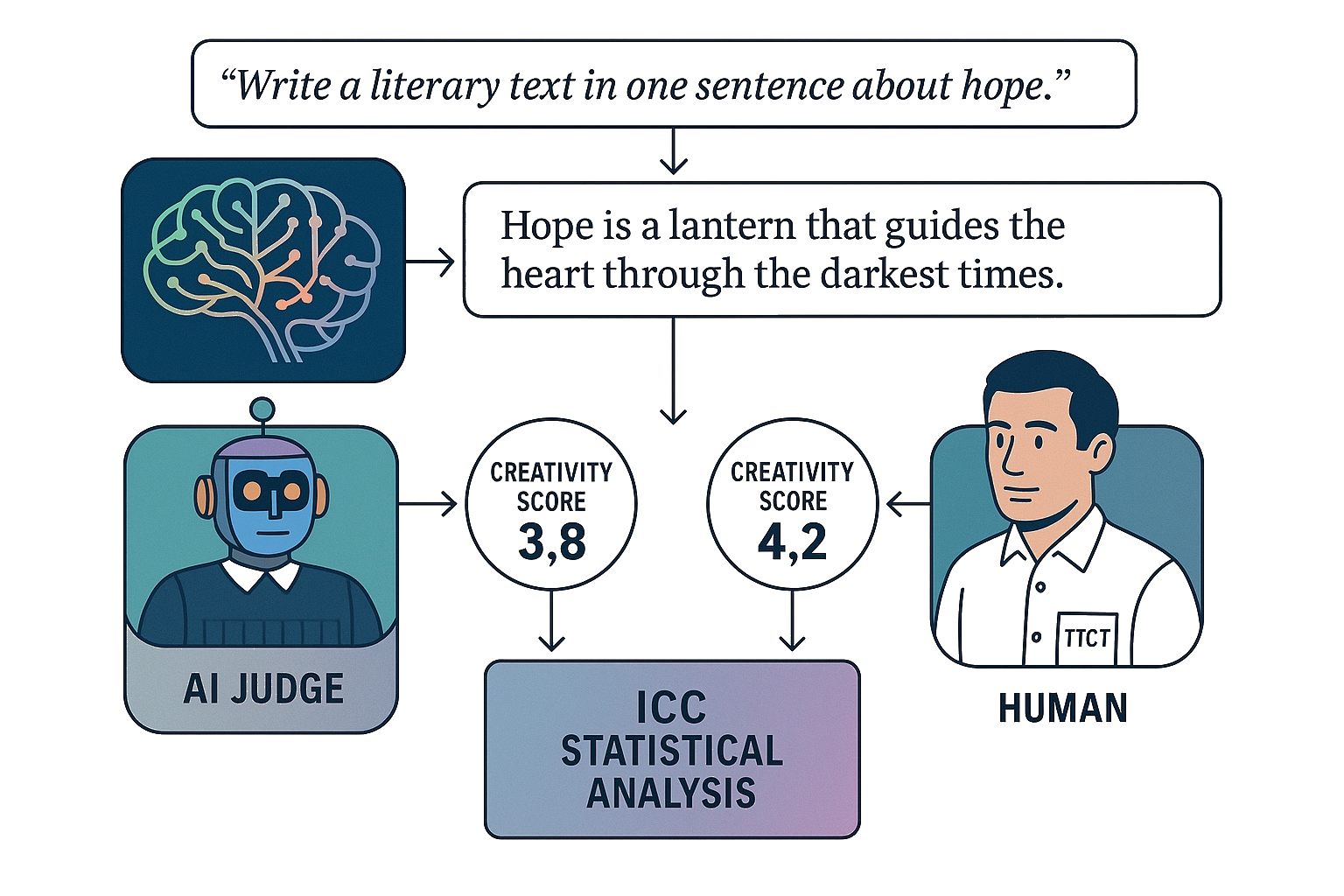}
    \caption{The overall flew of the evaluation framework and selection of the judge model.}
    \label{fig:flow}
\end{figure}

Given its stronger alignment with human judgments and its ability to capture culturally relevant stylistic nuances, we select Claude 3.7 Sonnet as the final LLM judge. This choice ensures that the automated evaluation framework remains consistent with human intuition and culturally grounded criteria. The overall evaluation process is summarized in Figure~\ref{fig:flow}.

\subsection{Prompting Strategy}
\label{sec:prompting}

To generate model outputs for evaluation, we adopt a consistent prompting strategy across all systems. Each model is instructed to produce 100 single-sentence literary texts for each of five themes—\textit{love}, \textit{longing}, \textit{friendship}, \textit{hope}, and \textit{despair}—corresponding to the topics selected for human annotations (Section~\ref{sec:human_set}). To simulate spontaneous human-like text production, we use a zero-shot prompt in Persian of the form: 
\textit{``Write a literary text in one sentence about \{Topic\}''}.

This prompt reflects how native speakers intuitively generate literary expressions: concise, topic-driven, and stylistically rich. All outputs are generated with a temperature setting of 1 to encourage creativity and variability while maintaining coherence. The resulting texts are then evaluated using the framework described in Section~\ref{sec:judge}.

Also in rare cases where a model produces duplicate sentences for different prompts (observed for DeepSeek-R1), we replace the repeated outputs with alternative, non-repetitive generations from the same model to preserve diversity in the evaluation.

\section{Experiments}
Here, we evaluate the creativity of Persian literary texts generated by various LLMs. We compare model performance, analyze the use of key rhetorical devices—specifically simile, metaphor, antithesis, and hyperbole—and assess the extent to which generated texts conform to Persian literary norms.

\subsection{Setup}
\label{sec:setup}
We use six LLMs to generate creative Persian literary texts: GPT-3.5-Turbo \cite{openai2023gpt35}, GPT-4.1 \cite{openai2025gpt41}, DeepSeek-V3-0324-671B \cite{deepseek2025v30324}, Gemma-3-27B-Instruct \cite{gemma2025instruct}, Qwen2.5-VL-32B-Instruct \cite{qwen2025vl32b}, and Deepseek-R1-671B \cite{deepseek2025r1}. These models were selected for their strong performance in generative and instruction-following tasks, representing a mix of proprietary and open-source systems with varying capabilities in multilingual and creative text generation.



\definecolor{headerblue}{RGB}{100, 149, 237}
\definecolor{rowgray}{gray}{0.95}

\begin{table}[t]
\centering
\renewcommand{\arraystretch}{1.5} 
\setlength{\arrayrulewidth}{0.4pt}
\setlength{\tabcolsep}{6pt}  
\resizebox{0.48\textwidth}{!}{
\begin{tabular}{>{\bfseries}l c c c c c}
\rowcolor{headerblue!40}
\textcolor{black}{\textbf{Model}} & 
\textcolor{black}{\textbf{Originality}} & 
\textcolor{black}{\textbf{Fluency}} & 
\textcolor{black}{\textbf{Flexibility}} & 
\textcolor{black}{\textbf{Elaboration}} & 
\textcolor{black}{\textbf{Creativity}} \\
\midrule

\rowcolor{white}
Gemma 3 & 0.035 & 0.006 & 0.044 & 0.023 & 0.010 \\

\rowcolor{rowgray}
Deepseek V3 & 0.050 & 0.030 & 0.044 & 0.012 & 0.015 \\

\rowcolor{white}
GPT-4.1 & 0.031 & 0.006 & 0.017 & 0.015 & 0.006 \\

\rowcolor{rowgray}
Qwen2.5 & 0.060 & 0.053 & 0.047 & 0.046 & 0.012 \\

\rowcolor{white}
GPT-3.5 & 0.021 & 0.065 & 0.078 & 0.062 & 0.021 \\

\rowcolor{rowgray}
Deepseek-R1 & 0.095 & 0.020 & 0.139 & 0.046 & 0.072 \\

\bottomrule
\end{tabular}%
}
\caption{\label{table:std_scores}
Standard deviations of scores for different models across evaluation criteria.}
\end{table}

\subsection{Comparing LLMs} \label{sec:compare_creativity}

To evaluate the creative potential of language models in a culturally grounded context, we assess their ability to generate Persian literary texts. The evaluation focuses on four core creativity dimensions applied to the texts (100 per model) generated in response to Persian literary prompts.

Each experiment is repeated three times to ensure reliability, and the reported values correspond to the average scores across runs. In addition, we compute the standard deviation of scores across the three runs, which provides a measure of stability for each model’s performance (see Table~\ref{table:std_scores}). The generally low standard deviations indicate that the evaluation is consistent and robust across repeated trials.

\begin{figure}[t]
    \centering
    \includegraphics[width=0.95\columnwidth]{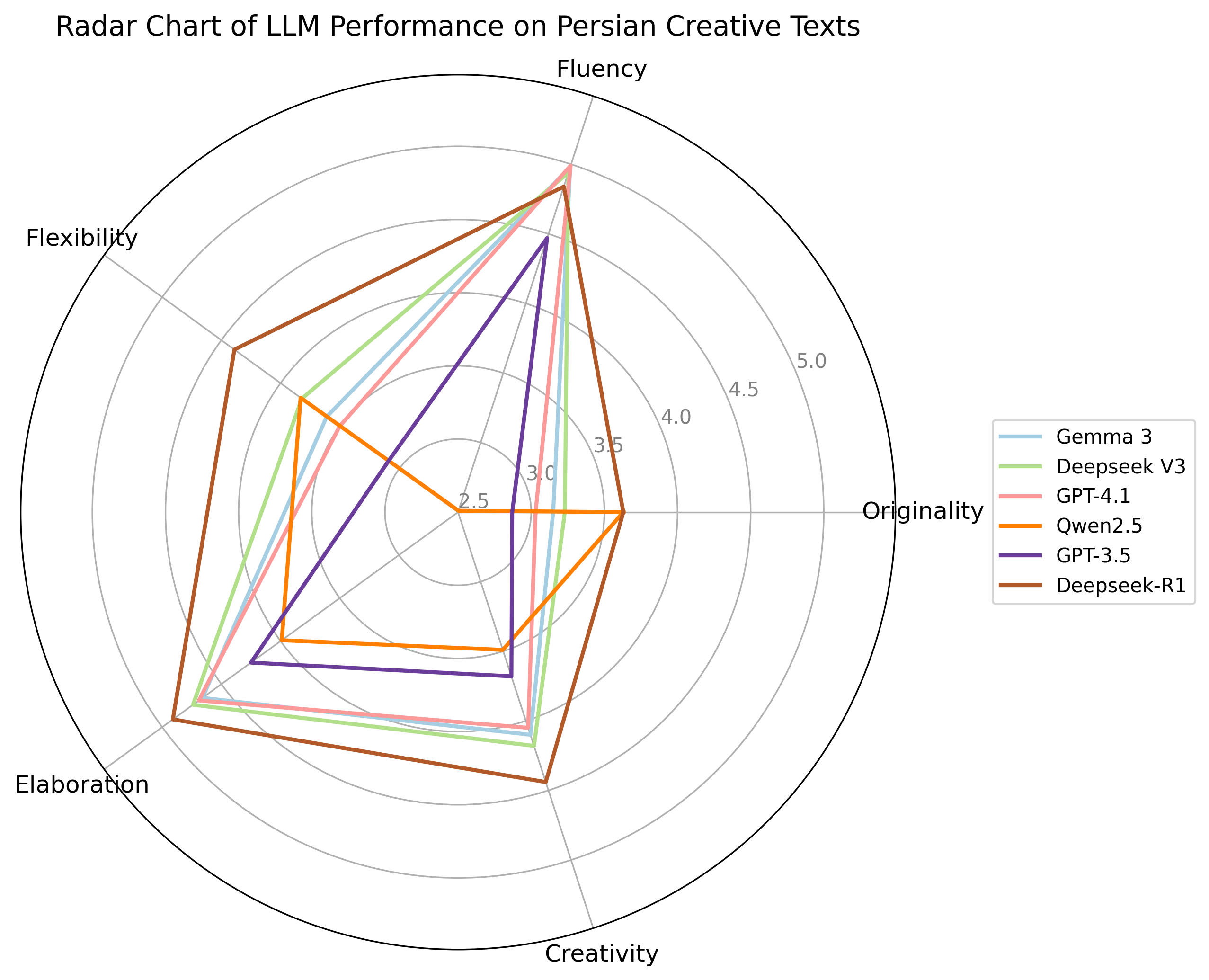}
    \caption{
    This figure compares creativity assessment scores across models. DeepSeek-R1 achieves the highest overall creativity score (4.44) and leads in flexibility (39) and elaboration (4.91). GPT-4.1 scores highest in fluency (4.99), while Qwen2.5-VL-32B-Instruct shows strong originality (3.63) but records the lowest fluency score (2.51), reflecting distinct performance patterns across different creativity dimensions.}
    \label{fig:creativity-comparison}
\end{figure}

Ratings are provided by the Claude 3.7 Sonnet model on a scale of 1 to 5 per dimension, and averages are calculated, including an overall creativity score (mean of all four dimensions), as shown in Figure~\ref{fig:creativity-comparison}.

DeepSeek-R1 achieves the highest overall creativity score among the evaluated models. Its strong performance in elaboration and flexibility indicates the model's ability to express diverse ideas, adopt multiple perspectives, expand on details, and evoke vivid sensory imagery. However, DeepSeek-R1 occasionally generates repetitive outputs. For instance, the sentence ``Love is the silent song of two hearts that, at the distance of a glance, breathe eternity in one breath'' appears three times on different prompts. To maintain diversity, we replace repeated outputs with other non-repetitive generations from the model.

One possible explanation for DeepSeek-R1's strong performance, particularly in elaboration and flexibility, is its reasoning-oriented training. The model is trained via reinforcement learning with objectives that promote structured thinking, including a ``think first, answer later'' approach \cite{deepseek2025r1}, which may implicitly support more detailed and diverse content generation.

DeepSeek-V3-0324 also demonstrates strong performance, particularly in elaboration and flexibility, suggesting that models in the DeepSeek family are capable of producing stylistically rich and imaginative literary texts with varied perspectives. Its low standard deviations across dimensions indicate stable and reliable performance.

\begin{figure}[ht]
    \centering
    \includegraphics[width=0.48\textwidth, trim=0 0 0 0, clip]{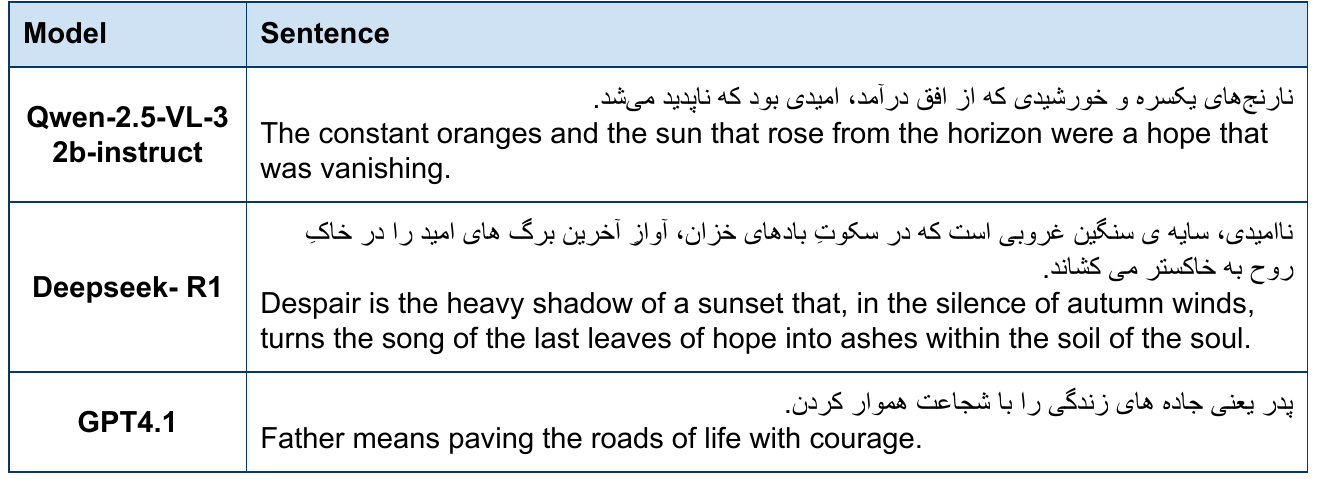}
    \caption{Sample sentences on despair from Qwen-2.5, DeepSeek-R1, and GPT-4.1, along with their English translations.}
    \label{fig:models_text}
\end{figure}

GPT-4.1 outperforms its predecessor, GPT-3.5, particularly in fluency and readability. Its outputs are grammatically correct and natural for Persian readers. However, it receives relatively low scores in originality, indicating that while its texts are coherent, they often rely on conventional expressions and lack inventive use of literary devices—though it still shows notable improvement over GPT-3.5. Compared to DeepSeek models, GPT-4.1 exhibits more consistent behavior, with smaller variance across runs.

In contrast, Qwen2.5-VL-32B-Instruct achieves higher originality scores but performs poorly in fluency. Its outputs are more novel and less clichéd, yet occasionally lack clarity and readability for Persian speakers. Additionally, the model scores lower in elaboration; while it introduces unique ideas, it struggles to create vivid imagery or convey emotional depth (e.g., love, sadness). The relatively larger standard deviations for Qwen2.5 in originality and fluency confirm this variability in creative performance.

These findings underscore the importance of evaluating creative text generation across multiple dimensions. Selecting models based on specific creativity criteria is essential for literary applications that require both stylistic authenticity and cultural nuance.

\section{Analysis}
\subsection{Word Frequency Analysis}

While Table~\ref{table:frequent_words_hope} does not reveal strong stylistic distinctions across all models, it suggests possible lexical similarities within model families. GPT-3.5 and GPT-4.1, for example, often rely on similar metaphorical terms such as hope, light, darkness, and heart. This may indicate that, despite architectural improvements, GPT-4.1 inherits certain lexical tendencies from GPT-3.5. The frequent use of binary oppositions like light/dark could reflect a preference for familiar, easily retrievable metaphors. Gemma-3-27B-IT exhibits a similar pattern, frequently reusing common symbolic contrasts, which may suggest a shared limitation in stylistic exploration.

A comparable trend is observed between DeepSeek-R1 and DeepSeek-V3-0324, which often use overlapping terms such as dark, night, and sound. These similarities may arise from shared training data, decoding strategies, or model architecture. While these terms are not inherently uncreative, their repeated use suggests that both models draw from a similar pool of literary expressions. In contrast, human-written texts display more varied and grounded imagery—e.g., sky and gaze—reflecting a more intuitive and emotionally nuanced approach to expression. These observations suggest that human creativity, even in short texts, tends to involve subtler and more diverse lexical choices than current LLMs typically produce.

\definecolor{headerblue}{RGB}{100, 149, 237}
\definecolor{rowgray}{gray}{0.95}
\definecolor{highlight}{RGB}{255, 255, 204}

\begin{table}[t]
\centering
\renewcommand{\arraystretch}{1.5}
\setlength{\arrayrulewidth}{0.4pt}
\setlength{\tabcolsep}{6pt}  
\resizebox{0.48\textwidth}{!}{
\begin{tabular}{>{\bfseries}l c c c c c}
\rowcolor{headerblue!40}
\textcolor{black}{\textbf{Text Source}} & 
\textcolor{black}{\textbf{1st Word}} & 
\textcolor{black}{\textbf{2nd Word}} & 
\textcolor{black}{\textbf{3rd Word}} & 
\textcolor{black}{\textbf{4th Word}} & 
\textcolor{black}{\textbf{5th Word}} \\
\midrule

\rowcolor{white}
Human & Hope (14) & Having (5) & Life (5) & Sky (3) & Gaze (3) \\

\rowcolor{rowgray}
GPT-3.5 & Hope (24) & Heart (16) & Light (7) & Darkness (5) & Bright (4) \\

\rowcolor{white}
GPT-4.1 & Hope (20) & Night (15) & Heart (11) & Dark (8) & Bright (6) \\

\rowcolor{rowgray}
Gemma-3-27B-Instruct & Hope (20) & Night (19) & Dark (17) & Light (17) & Star (12) \\

\rowcolor{white}
Qwen2.5-VL-32B-Instruct & Hope (23) & Light (9) & Heart (8) & Darkness (6) & Life (5) \\

\rowcolor{rowgray}
QwQ-32B & Hope (20) & Darkness (7) & Light (6) & Black (4) & Sky (4) \\

\rowcolor{white}
DeepSeek-V3-0324 & Hope (20) & Darkness (12) & Night (11) & Bird (9) & Sound (6) \\

\rowcolor{rowgray}
DeepSeek-R1 & Hope (19) & Dark (11) & Night (9) & Sunrise (8) & Sound (7) \\

\bottomrule
\end{tabular}%
}
\caption{\label{table:frequent_words_hope}
Top 5 most frequent words generated on the theme of hope across different text sources (frequency in parentheses).}
\end{table}

In this context, reasoning-oriented LLMs appear to positively influence the outputs. Specifically, the frequency of common and repetitive words decreases, while more creative and human-like terms emerge. For instance, in DeepSeek-R1, the word ``sunrise'' appears more frequently—a term that is more vivid and imaginative compared to simpler descriptors like ``dark'' or ``light''.

To further explore this effect, we also evaluated another reasoning-enhanced model, QWQ-32B, an improved version of Qwen. This model similarly shows an increased frequency of the word ``sky'', a term commonly found in human-generated texts. In both cases, the frequency of highly repeated words, such as ``hope'' and ``dark'', decreases relative to their corresponding base models (i.e., models from the same family without reasoning enhancements). Detailed analyses of word relationships, text similarity,and creativity metrics are left for future work.

\begin{figure}[t]  
    \centering
    \includegraphics[width=0.95\columnwidth]{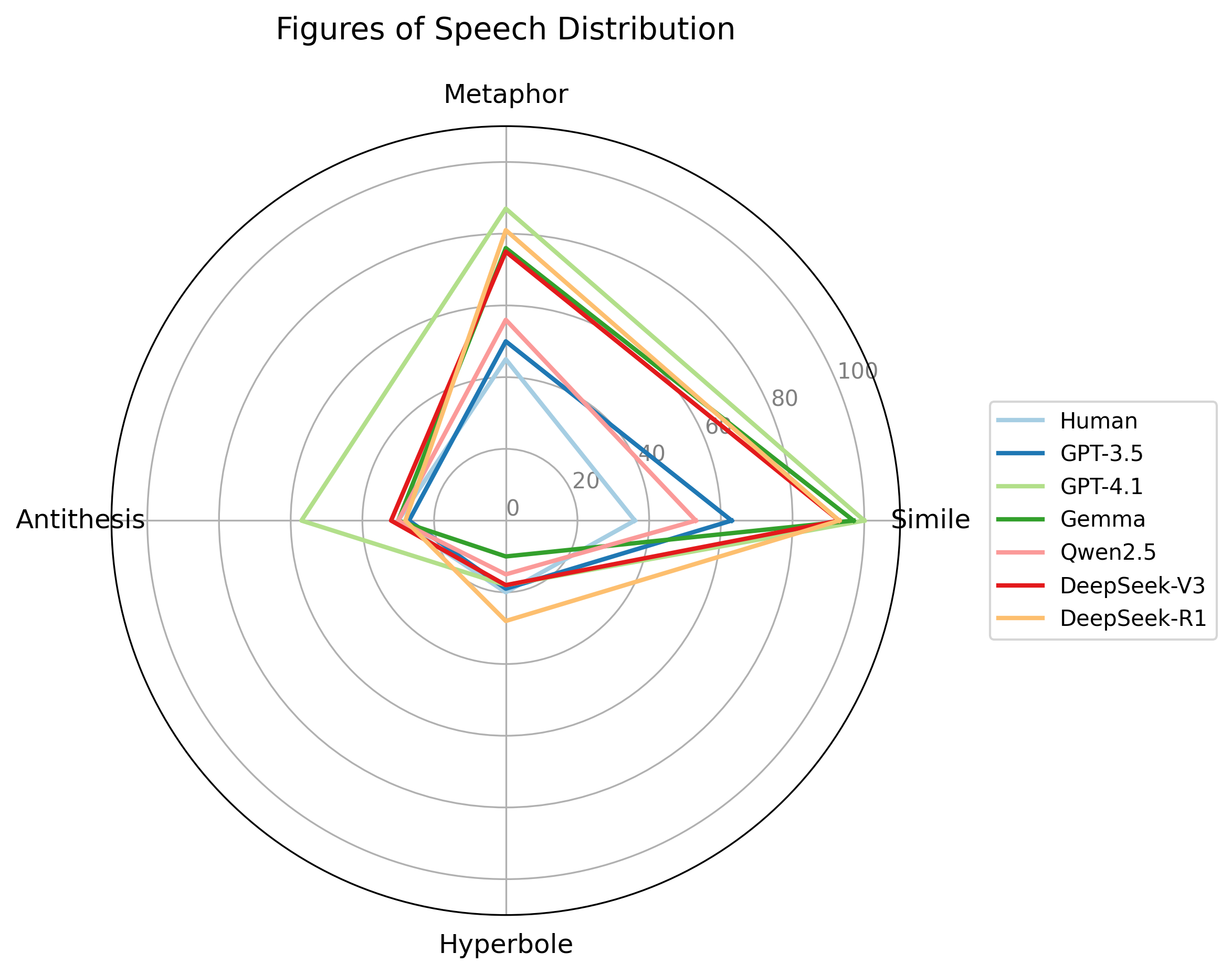}
    \caption{
    Count of figures of speech (simile, metaphor, antithesis, hyperbole) used in human-written and model-generated texts across different models.}
    \label{fig:figures_of_speech_comparison_counts}
\end{figure}

\subsection{Figure of Speech Analysis}
\label{sec:fsa}

To evaluate the stylistic richness and creative capacity of generated texts, we analyzed the use of four common figures of speech—simile, metaphor, antithesis, and hyperbole—comparing human-written texts with outputs from the six LLMs, as shown in Figure~\ref{fig:figures_of_speech_comparison_counts}.


Two human annotators independently labeled the rhetorical devices, achieving inter-annotator agreement rates of 80\% for \textit{Human texts} and 84\% for \textit{Model texts}. We initially intended to use LLMs as judges in the same manner as for text creativity. However, both Claude and GPT-4o exhibited inconsistencies in identifying figures of speech, particularly similes and metaphors. These models often confused the two devices and did not demonstrate a clear understanding of their distinctions. Consequently, we relied on human annotators to accurately identify the rhetorical devices in each text.

Human-written texts display a balanced use of literary devices—primarily metaphors and similes—reflecting natural stylistic coherence. Among the models, GPT-4.1 produces the highest number of rhetorical devices overall, with similes appearing in all 100 of its generated texts. This overreliance may partly explain its lower originality score. In contrast, Qwen2.5-VL-32B-Instruct demonstrates a more selective and varied use of rhetorical devices, aligning with its higher originality rating. This suggests that a deliberate and inventive deployment of literary elements can better enhance perceived creativity.

DeepSeek-R1, while employing metaphors and similes extensively, also shows a more balanced application of antithesis and hyperbole. Its effective use of hyperbole, in particular, likely contributes to its strong elaboration scores by enhancing vivid imagery and mental representation.

Overall, our analysis indicates that the sheer quantity of stylistic devices alone does not guarantee creativity. Rather, the type, balance, and nuanced application of figures of speech play a crucial role in shaping the originality, fluency, flexibility, and elaboration of generated texts.

\section{Conclusion}

In this work, we introduce a novel framework for evaluating creativity in LLMs within the context of Persian literary text generation. Building on the TTCT, we propose a culturally adapted evaluation scheme that captures four core dimensions—originality, fluency, flexibility, and elaboration—which can be applied to any language. Our analysis shows that no single model performs well across all dimensions of creativity. Instead, different models exhibit varying strengths in different aspects: for example, DeepSeek-R1 and GPT-4.1 demonstrate high expressive richness, whereas others, such as Qwen2.5, generate more concise yet culturally resonant outputs. This highlights that creative ability in LLMs is distributed unevenly across dimensions rather than concentrated in a single model.

A key outcome of our analysis is that models tend to follow learned patterns rather than demonstrating genuinely diverse creativity. Their use of rhetorical devices skews heavily toward simile and metaphor, with limited balance across other devices such as antithesis and hyperbole. Human texts, in contrast, often integrate multiple figures of speech within a single sentence, resulting in richer and more creative expression. Moreover, when we attempt to use LLMs as judges for labeling rhetorical devices, they struggle—particularly in distinguishing between metaphor and simile—highlighting current limitations in nuanced literary understanding.

We also curate a unique dataset, CPers, comprising 4,371 single-sentence literary texts spanning diverse topics and emotions. This dataset is the first of its kind in Persian, and, to the best of our knowledge, no comparable dataset exists in English. We also release our human-annotated subset, including 100 human-written and 600 model-generated texts, with annotations covering both creativity scores and the figures of speech employed in each text.

Taken together, our findings indicate that LLMs serve as useful tools for Persian literary text generation, but expectations remain modest: their creativity does not yet parallel human-level diversity and literary nuance. These results emphasize the need for culturally grounded evaluation in multilingual NLP, particularly for low-resource, high-context languages.

Future work explores increasing the number of annotators, improving the judge model, examining more diverse topics, testing diverse prompting strategies, and conducting cross-lingual comparisons to assess the adaptability of LLM creativity across cultures.

\section*{Limitations}

While our framework provides a structured and culturally grounded approach to evaluating creativity in Persian literary text generation, it is not without limitations. The evaluation questions were designed and scored by the authors, who—while fluent in the language and familiar with literary conventions—are not formally trained in psychology or Persian literary studies. Future work could benefit from interdisciplinary collaboration to refine both the criteria and the evaluation process.

Our analysis focused on 100 samples and five themes, offering a practical but narrow window into the broader dataset. Creativity, however, often unfolds more vividly across longer narratives and diverse emotional contexts. Evaluating paragraph-level or multi-sentence outputs may uncover richer stylistic patterns and deeper coherence that go unnoticed at the sentence level.

We also restricted our prompting to a zero-shot setup. Exploring other prompting strategies—such as few-shot, chain-of-thought, or instruction prompting—could help reveal how different models respond to varying task formulations, and whether prompt design can shape creativity in meaningful ways.

Moreover, while our focus on Persian fills a critical gap, it leaves open the question of how these models perform across languages. A cross-lingual comparison would shed light on whether the observed creative behaviors are language-dependent or model-intrinsic, and could further reveal how cultural and linguistic structure shape creative expression.



\bibliography{acl_latex}

\appendix

\section{Dataset Construction}
\label{sec:appendix}

The \textit{CPers} dataset, introduced in Section~\ref{sec:dataset}, contains 4,371 short literary-style Persian texts collected from a variety of online sources. These texts reflect a wide range of cultural and emotional themes and are primarily written by native Persian speakers.

\subsection*{Topics Covered}

The dataset covers 20 culturally significant themes, including love, mother, father, longing, birthday, boy, girl, Yalda Night (an Iranian celebration marking the longest night of the year), friendship, Nowruz (the Iranian New Year), autumn, winter, spring, summer, despair, sorrow, life, separation, hope, and kindness.

\subsection*{Collection Process}
The texts are gathered from publicly available website and blogs featuring Persian literary and emotional content. We focuse on collecting relatively short texts, typically one sentence or a few lines, suitable for sentence-level creativity evaluation.
All data instances have been reviewed by humans to ensure they do not contain any personal information or offensive content.

\subsection*{Data Usage and Disclaimer} \label{appendix:data}
The data from online resources used to create the dataset is anonymized and publicly available. The \textit{CPers} dataset is intended for research purposes.

\subsection*{Source Attribution}
Texts are sourced from a range of publicly available platforms.\footnote{Example sources include: 

\url{https://digipostal.ir}, \url{https://www.beytoote.com}, \url{https://roozaneh.net}, \url{https://vista.ir}, \url{https://shereno.com}, \url{https://salamdonya.com}, \url{https://chishi.ir}, \url{https://www.delgarm.com}, \url{https://diamag.ir}, \url{https://fararu.com}, \url{http://www.coca.ir}, \url{https://www.alamto.com}, \url{https://setare.com}, \url{https://wikimatn.com}, \url{https://robinarose.com}, \url{https://www.tasvirezendegi.com}, \url{https://www.talab.org}, \url{https://delbaraneh.com}, \url{https://www.bishtarazyek.com}, \url{https://topnaz.com}, 
\url{https://magerta.ir}, \url{https://namnak.com}}

\section{Confusion Matrices}\label{appendix:conf}
\begin{figure*}[h]
    \centering
    \subfloat[ ]{
    \includegraphics[height=.35\textwidth,trim=10mm 0 10mm 0]{fig/org_hh_crop.png}
    
    }\hspace{5mm}
    \subfloat[ ]{
	\includegraphics[height=.35\textwidth,trim=10mm 0 10mm 0]{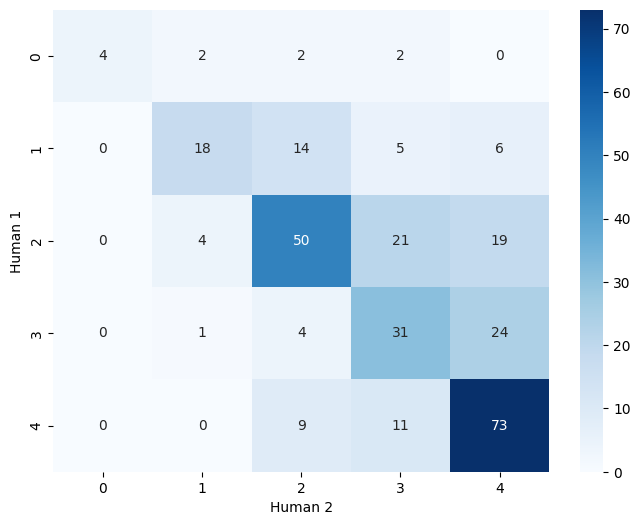}
	}
	\vspace{-3mm}
    \caption{Confusion matrices showing inter-rater agreement between humans on Model texts: (a) Originality criteria, and (b) Fluency criteria.
    }
    \label{fig:6}
    \vspace{-3mm}
\end{figure*}

\begin{figure*}[h]
    \centering
    \subfloat[]{
    \includegraphics[height=.35\textwidth,trim=10mm 0 10mm 0]{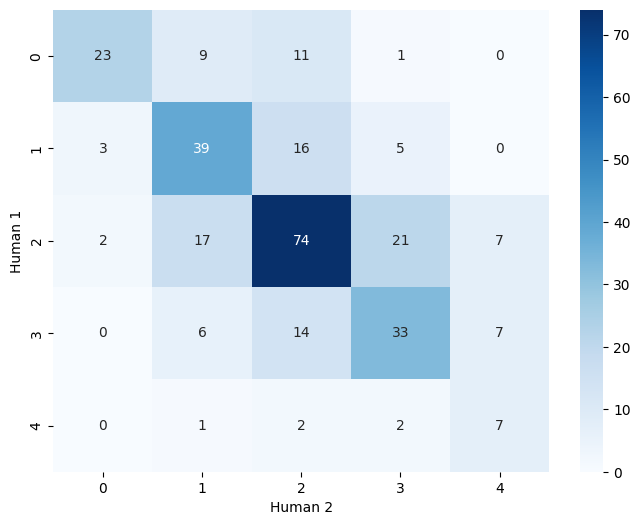}
    
    }\hspace{5mm}
    \subfloat[ ]{
	\includegraphics[height=.35\textwidth,trim=10mm 0 10mm 0]{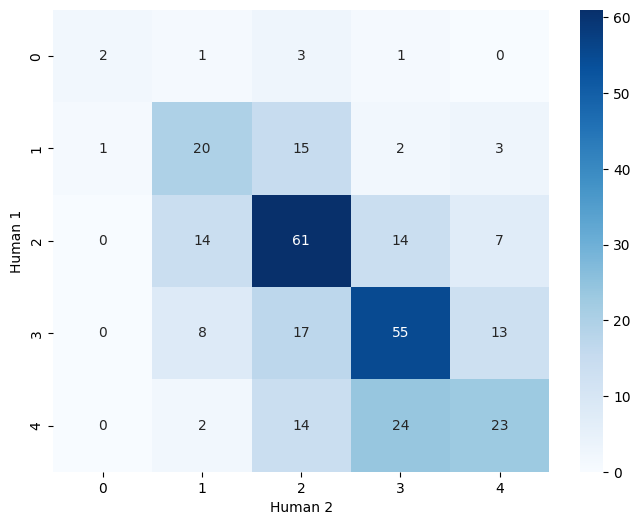}
	}
	\vspace{-3mm}
    \caption{Confusion matrices showing inter-rater agreement humans on Model texts: (a) Flexibility criteria, and (b) Elaboration criteria.
    }
    \label{fig:7}
    \vspace{-3mm}
\end{figure*}

\begin{figure*}[h]
    \centering
    \subfloat[]{
    \includegraphics[height=.35\textwidth,trim=10mm 0 10mm 0]{fig/human_org.png}
    
    }\hspace{5mm}
    \subfloat[ ]{
	\includegraphics[height=.35\textwidth,trim=10mm 0 10mm 0]{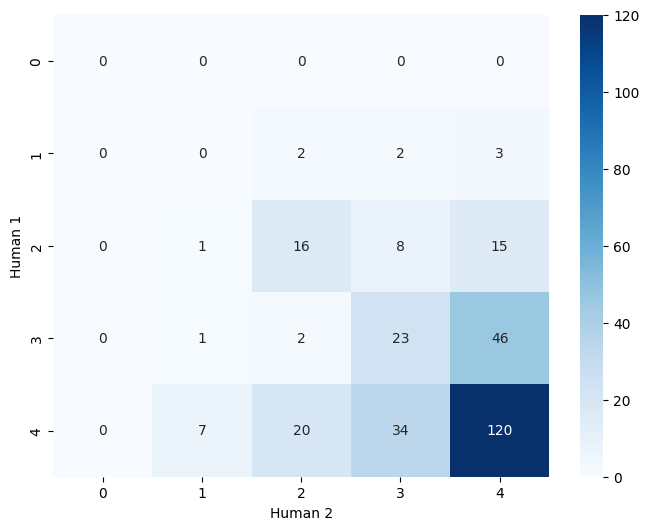}
	}
	\vspace{-3mm}
    \caption{Confusion matrices showing inter-rater agreement between humans on Human texts: (a) Originality criteria, and (b) Fluency criteria.
    }
    \label{fig:8}
    \vspace{-3mm}
\end{figure*}

\begin{figure*}[h]
    \centering
    \subfloat[ ]{
    \includegraphics[height=.35\textwidth,trim=10mm 0 10mm 0]{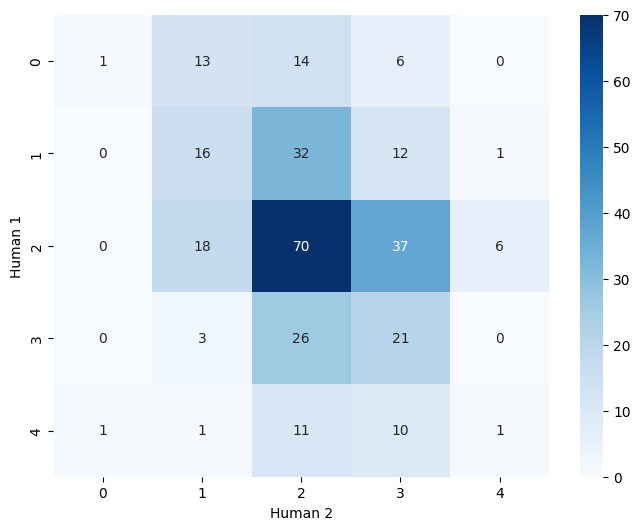}
    
    }\hspace{5mm}
    \subfloat[]{
	\includegraphics[height=.35\textwidth,trim=10mm 0 10mm 0]{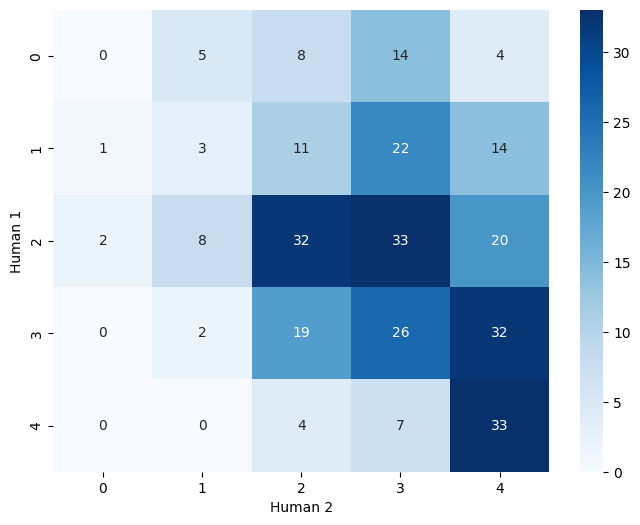}
	}
	\vspace{-3mm}
    \caption{Confusion matrices showing inter-rater agreement between humans on Human texts: (a) Flexibility criteria, and (b) Elaboration criteria.
    }
    \label{fig:9}
    \vspace{-3mm}
\end{figure*}

Confusion matrix to show agreement between human annotators across all criteria and text types are presented in Figures \ref{fig:6}, \ref{fig:7}, \ref{fig:8}, and \ref{fig:9}.

\end{document}